\documentclass[sigconf]{acmart}
\usepackage{tabularx}

\copyrightyear{2026}
\acmYear{2026}
\setcopyright{cc}
\setcctype{by}
\acmConference[SIGGRAPH Posters '26]{Special Interest Group on Computer Graphics and Interactive Techniques Conference Posters}{July 19--23, 2026}{Los Angeles, CA, USA}
\acmBooktitle{Special Interest Group on Computer Graphics and Interactive Techniques Conference Posters (SIGGRAPH Posters '26), July 19--23, 2026, Los Angeles, CA, USA}
\acmDOI{10.1145/3799825.3818709}
\acmISBN{979-8-4007-2548-7/2026/07}

\begin{document}

\title{Appearance-Invariant Detection of Suggestive Motion via Laban Movement Descriptors}

\author{Jaehoon Ahn}
\affiliation{%
  \institution{Sogang University}
  \city{Seoul}
  \country{South Korea}
}
\email{jahn@sogang.ac.kr}

\author{Jeonghan Kong}
\affiliation{%
  \institution{Sogang University}
  \city{Seoul}
  \country{South Korea}
}
\email{k7kong@sogang.ac.kr}

\author{Moon-Ryul Jung}
\authornote{Corresponding author}
\affiliation{%
  \institution{Sogang University}
  \city{Seoul}
  \country{South Korea}
}
\email{moon@sogang.ac.kr}

\date{April 2026}

\begin{abstract}
Content moderation in online multiplayer 3D virtual environments is increasingly automated, yet detection has focused on images, video, and audio, leaving suggestive motion a blind spot. We present a motion-only classification pipeline that detects suggestive and explicit movement from SMPL skeleton trajectories using Laban Movement Analysis (LMA) descriptors. On a dataset spanning everyday, artistic, suggestive, and explicit movement (17+ hours of video), a logistic regression trained on 61-feature LMA descriptors reaches 68\% binary SFW/NSFW accuracy (70\% random forest) under a leak-free evaluation protocol. At this level, our descriptor performs comparably to a learned video model trained on the same motion re-rendered as appearance-free video, a gray figure with no clothing, skin, or scene. The indirectness (tortuosity) of each joint's trajectory, measured as the ratio of the joint's path length to its net displacement, peaks at the suggestive tier, showing that the Direct-to-Indirect polarity of Laban's Space factor provides an interpretable marker of the shift from functional to suggestive motion. Ultimately, Laban-based kinematic descriptors offer a lightweight, interpretable approach to suggestive-motion detection: every decision decomposes into named, theory-grounded features. Because the classifier operates on pose trajectories alone, moderation can run directly on avatar poses in virtual environments, with no appearance data.
\end{abstract}

\maketitle

\section{Introduction}
3D social platforms such as VRChat and Roblox allow users to animate avatars with arbitrary appearance. Conventional not-safe-for-work (NSFW) detection relies on pixel-level features; in practice, classifiers learn to flag revealing clothing, skin exposure, and body shape rather than the motion itself. We confirmed this directly: Qwen3-VL \cite{bai2025qwen3}, a state-of-the-art multimodal LLM, cannot classify suggestive motion from skeleton-only visualizations. A moderation system for motion suggestiveness must therefore operate on motion rather than appearance.

We observe that choreographic theory, specifically Rudolf Laban's \emph{Effort} framework~\cite{laban1971mastery}, provides an interpretable vocabulary for quantifying qualitative shifts in motion. To test whether LMA descriptors can empirically separate safe from unsafe motion, we extract 3D SMPL \cite{loper2023smpl} skeletons (camera frame) from video using WHAM~\cite{shin2024wham} and compute an LMA descriptor, then classify motion fragments along a suggestiveness taxonomy.\footnote{Code, the 61-feature data, and a reproduction notebook are available at \texttt{https://github.com/zaiisao/suggestive-motion-lma}.} Using solely the LMA features---no pixels, clothing, or body shape---a logistic regression separates safe from unsafe (SFW/NSFW) motion at ${\approx}68\%$ accuracy under a leak-free, clip-length-controlled evaluation, matching a learned video model fine-tuned on the same motion with appearance removed. We evaluate on web and academic video from Kinetics-700 \cite{kay2017kinetics, carreira2019short, smaira2020short}, TikTok/YouTube, and NPDI \cite{avila2013pooling}. The descriptors require only a fixed set of 3D key-joint positions and are not inherently specific to SMPL.

Our contributions are:
\begin{enumerate}
  \item We apply LMA descriptors, originally validated for dance style recognition, to the novel task of classifying motion along a suggestiveness gradient, using WHAM to jointly estimate floor contact and the 3D skeleton.
  \item Empirical evidence that all four Laban Effort factors carry discriminative signal, with different factors dominating at different granularities, validating the LMA framework computationally.
  \item Baseline binary (SFW/NSFW) and three-way (everyday, suggestive, and explicit) classifiers that perform comparably to a learned video model trained on the same appearance-invariant motion, establishing a lower bound for motion-only moderation, with a four-way ablation probing the artistic--suggestive boundary.
\end{enumerate}

\section{Method}
 
\begin{figure*}[t]
  \centering
  \includegraphics[width=0.9\textwidth]{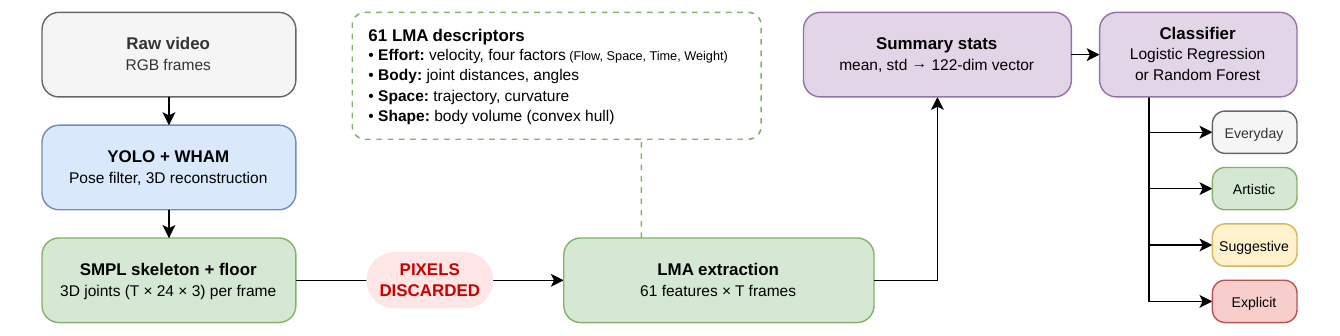}
  \caption{Motion classification pipeline.  Raw video is processed by
    WHAM to produce per-frame SMPL parameters; 61 LMA descriptors
    are computed per frame and summarized to a 122-dim vector
    (mean, std) per fragment.  At no point does the classifier
    access pixel data.}
  \label{fig:pipeline}
\end{figure*}
 
\paragraph{Data.}
We construct a four-tier dataset of motion fragments:

\begin{itemize}
    \item \textbf{Tier 0} (everyday: walking, sitting, dining) from Kinetics-700 (19 classes, 515 videos, 1,027 fragments)
    \item \textbf{Tier 1} (artistic: breakdancing, ballet, capoeira, gymnastics) from Kinetics-700 (16 classes, 412 videos, 1,889 fragments)
    \item \textbf{Tier 2} (suggestive: twerk, reggaeton/perreo, sensual fancam, streamer dance cover, chair dance, heels, belly dance) crawled from YouTube (7 categories) and TikTok; see Appendix~\ref{app:tier2} (108 videos, 1,715 fragments)
    \item \textbf{Tier 3} (explicit) from the NPDI academic corpus (238 videos, 2,620 fragments)
\end{itemize}

\noindent For the binary task, 480 Tier-1 fragments join Tier 0 to form the safe class; Tier 1 is otherwise excluded from the three-way and forms its own class in the four-way.
 
\paragraph{Skeleton extraction.}
Each video passes through YOLO11x-pose filtering \cite{Jocher_Ultralytics_YOLO_2023} (retaining frames with $\geq$10 keypoints above confidence $0.4$ and mean keypoint confidence $\geq$0.5), continuous segment extraction ($\geq$3\,s), and WHAM inference~\cite{shin2024wham} to produce per-frame SMPL joint positions in the camera frame.

\paragraph{LMA features.}
We compute 61 descriptors per frame\footnote{\citeauthor{turab2025dance} use 55 LMA features but publish no feature list or code; a literal reading of their descriptor yields 61, so we retain all 61 rather than drop six to force the count.} grouped into Laban's four components: \emph{Body} (inter-joint distances, angles, and movement initiation), \emph{Effort} (per-joint velocity and the Flow, Space/Directness, Time, and Weight factors), \emph{Space} (limb dispersion, trajectory path, and curvature), and \emph{Shape} (convex-hull body volume). Each fragment's $(T \times 61)$ matrix is collapsed to a 122-dimensional vector via per-feature mean and standard deviation (Fig.~\ref{fig:pipeline}).
 
\paragraph{Classification.}
We train logistic regression and random forest classifiers on class-balanced samples under five-fold cross-validation. A single source video can yield multiple fragments when several performers are present, or even when a dancer passes behind another and reappears, a situation common in dance fancams. We therefore group the folds by source video, keeping all of a video's fragments in one fold so the classifier cannot match a performer or scene seen in training to the test set. The three- and four-way tasks use 1{,}027 fragments per class. The binary task uses 1{,}507 per class: the safe class is everyday plus artistic motion (1{,}027 T0 and 480 T1), and the unsafe class is sampled from the suggestive and explicit pool (${\approx}40\%$ T2, ${\approx}60\%$ T3).

\section{Results}
 


\begin{table}[t]
  \centering
  \small
  \caption{Appearance-invariant motion classification on identical fragments and source-video-grouped folds. Hand-crafted LMA descriptors vs.\ a learned video model on appearance-free re-renderings of the same motion (a uniform gray figure on a blank background); the large drop is the clip-length confound, after which the models are at parity.}
  \label{tab:videomae}
  \begin{tabular}{lccc}
    \toprule
    & Binary & Three-way & Four-way \\
    \midrule
    \multicolumn{4}{l}{\textit{Source-video-grouped cross-validation}} \\
    \quad LMA (logistic regression) & 0.778 & 0.706 & 0.576 \\
    \quad LMA (random forest)       & \textbf{0.809} & \textbf{0.708} & \textbf{0.578} \\
    \quad VideoMAE (appearance-free)           & 0.724 & 0.664 & 0.557 \\
    \midrule
    \multicolumn{4}{l}{\textit{+ length-matched (clip-length confound removed)}} \\
    \quad LMA (logistic regression) & 0.678 & 0.635 & 0.520 \\
    \quad LMA (random forest)       & \textbf{0.704} & \textbf{0.638} & 0.530 \\
    \quad VideoMAE (appearance-free)           & 0.694 & 0.628 & \textbf{0.532} \\
    \bottomrule
  \end{tabular}
\end{table}

\paragraph{Classification accuracy.}
Under source-video-grouped five-fold cross-validation---every fragment of a source video confined to a single fold, so no near-duplicate motion is shared between train and test---logistic regression over the 61-feature LMA descriptor reaches \textbf{0.78} binary SFW/NSFW accuracy (chance 0.50) and \textbf{0.71} three-way (everyday/suggestive/explicit; chance 0.33); a random forest reaches \textbf{0.81} and \textbf{0.71} for binary and three-way respectively.

\paragraph{Controlling for clip length.}
One artifact qualifies the binary number specifically. NSFW fragments, which are dominated by crowd- and occlusion-heavy Tier-3 footage, are shorter on average than SFW fragments, and short clips yield degraded, partial-window WHAM reconstructions. Thus, a classifier can exploit duration as a shortcut: clip length \emph{alone} separates the two binary classes at 0.62. We remove this shortcut by binning fragments into ten length deciles and subsampling each class to equal counts per bin, resulting in identical clip-length distributions. On the length-matched data the accuracies settle to \textbf{0.68} binary and \textbf{0.64} three-way (random forest \textbf{0.70} and \textbf{0.64}). 

\paragraph{An interpretable descriptor matches a learned video model.}
To ask whether a learned model extracts motion signal the hand-crafted descriptor misses, we re-render every fragment as an appearance-free video in which the reconstructed motion is played back by a featureless gray figure on a blank background, with clothing, skin, body shape, and scene removed. We use these rendered videos to fine-tune VideoMAE~\cite{tong2022videomae} on the \emph{identical} fragments and source-video-grouped folds. Once the clip-length confound is removed, the two are indistinguishable across binary, three-way, and four-way---within ${\approx}1$\,pp on every task (Table~\ref{tab:videomae}, lower panel). On the raw grouped data the LMA models lead by up to 8.5\,pp, but that margin is the confound, not motion signal. An interpretable 61-D descriptor thus matches a finetuned video transformer on the same appearance-free motion, at no cost to auditability.

\paragraph{Artistic ablation.}
Reintroducing the artistic tier defines a four-way task (logistic regression 0.58, chance 0.25), in which the artistic--suggestive boundary (T1$\leftrightarrow$T2) is by far the hardest: 22\% of artistic fragments are predicted as suggestive, and 24\% of suggestive as artistic. Both involve continuous high-energy dance and share a kinematic envelope. The everyday and explicit extremes recover cleanly (T0=\,68\%, T3\,=\,67\% recall), and a counterintuitive finding is that explicit motion confuses more with everyday (17\%) than with suggestive (8\%); static explicit content is kinematically close to ordinary posture. Because artistic motion is not cleanly separable from suggestive, we report four-way only as a diagnostic.
 
\begin{figure}[t]
  \centering
  \includegraphics[width=0.9\columnwidth]{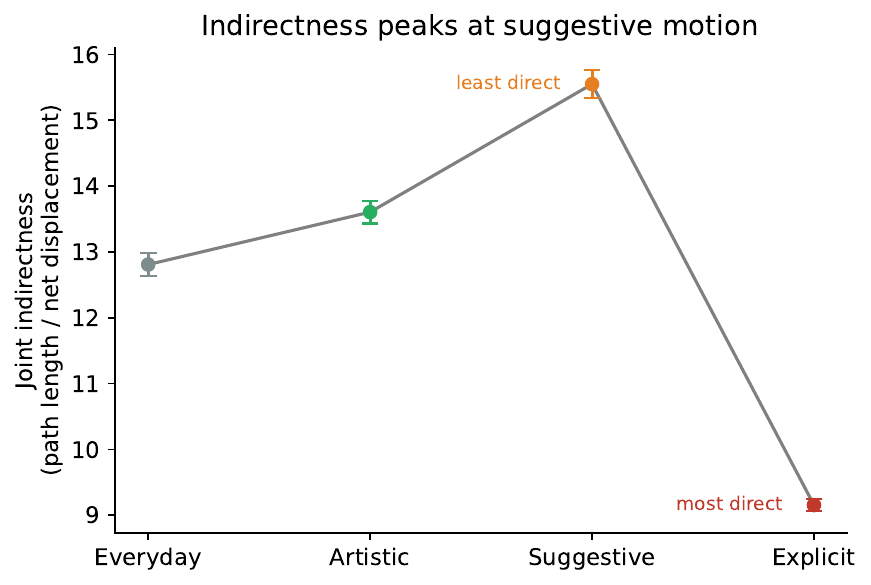}
  \caption{Mean joint indirectness (tortuosity: trajectory path length over net displacement) per tier. The profile is non-monotonic: suggestive motion is the most indirect while explicit motion is the most direct. Error bars: $\pm$1 s.e.m.}
  \label{fig:directness}
\end{figure}
 
\paragraph{What the descriptor keys on.}
Different Effort factors lead at different granularities (Kruskal--Wallis $H$): Time (acceleration) and Flow (jerk) are the strongest discriminators in the three- and four-way splits, Time at rank~3. Directness (the Space factor of Effort) is non-monotonic (Fig.~\ref{fig:directness}): it peaks at suggestive motion (T2), whose joints trace recirculating, indirect paths that \emph{display} rather than \emph{navigate}, and is lowest at explicit motion (T3), the most direct of all. By utilizing LMA's Effort Space vocabulary, we can interpret this result functionally: the shift from athletic to suggestive motion aligns with the Direct-to-Indirect polarity~\cite{laban1971mastery}, but only per tier: in the safe/unsafe binary it washes out, since explicit motion is itself direct and dominates the unsafe class. 
 

\section{Conclusion}

\paragraph{Implications.}
Shallow, interpretable classifiers over LMA features are as effective at flagging suggestive motion as a fine-tuned deep video transformer on the same appearance-free motion---from skeleton alone, with no access to appearance. The interpretability of LMA features makes the system auditable: each feature has a name, a choreographic meaning, and a century of dance-theory context~\cite{chi2000emote, samadani2013laban}.

\paragraph{Limitations.}
Two caveats remain. First, each tier is drawn from a single source, so dataset and suggestiveness label are largely collinear; we cannot cleanly separate motion signal from source-specific artifacts, and address only the largest of them, clip length. Second, grouping folds by source video prevents within-video leakage, but a performer appearing in several different videos can still recur across folds, the dancer-identity confound familiar from AIST++, which we cannot remove without per-performer labels.

\paragraph{Future work.}
The four-way baseline (${\approx}58\%$) sets the floor that temporally-attentive models must beat. We plan to fine-tune MotionBERT~\cite{zhu2023motionbert} on the raw $(T\times61)$ sequences and fuse it with the interpretable LMA descriptors in a two-stream model, keeping LMA's auditability while capturing the temporal dynamics our per-fragment summaries discard. Collecting each suggestiveness level from multiple platforms would further decouple motion from source, and we aim to generalize from reconstructed skeletons to metaverse avatar rigs.

\bibliographystyle{ACM-Reference-Format}
\bibliography{references}

\appendix

\section{Pipeline implementation details}
We give the settings needed to reproduce the pipeline without reference to our code.

\paragraph{Pose and fragments.}
We run WHAM~\cite{shin2024wham} per detected person. Within WHAM, person boxes are detected with YOLO26x (confidence $0.4$) and 2D keypoints with ViTPose++ (base, $256{\times}192$, UDP)~\cite{xu2023ViTPose++}, with cross-frame identity association by object--keypoint similarity (OKS, threshold $0.25$); a track is kept only if it spans at least $\max(30,\lfloor 2.5\,\mathrm{fps}\rfloor)$ frames ($\approx 75$ at $30$\,fps). We recover the 24 SMPL joints in the \emph{camera} frame by applying the SMPL 24-joint regressor to the posed SMPL vertices, $J = \mathcal{R}\,V$, where $\mathcal{R}$ is the regressor and $V$ the vertices. (We regress joints from vertices rather than reading WHAM's native 31-joint COCO+SPIN output, whose first 24 entries are \emph{not} the SMPL joints.) Upstream of WHAM, we keep frames where YOLO11x-pose~\cite{Jocher_Ultralytics_YOLO_2023} reports $\geq 10$ keypoints above confidence $0.4$ with mean keypoint confidence $\geq 0.5$, and extract continuously tracked segments of $\geq 3$\,s. A \emph{fragment} is one tracked person within one continuous segment; fragments shorter than the descriptor window ($55$ frames) are discarded (in practice the track-length gate above binds first).

\paragraph{The LMA features.}
Features are computed per frame over a sliding window of $W=55$ frames on the six key joints---head, pelvis, left/right wrist, left/right ankle---following Turab et al.~\cite{turab2025dance}, and group into four Laban categories summing to 61 (Table~\ref{tab:features}). Per joint we take the mean velocity magnitude (Effort) and the four Effort factors: \emph{Weight} as mean kinetic energy $\tfrac12\lVert v\rVert^2$; \emph{Time} as mean acceleration magnitude $\lVert a\rVert$; \emph{Flow} as mean jerk magnitude $\lVert \dot{a} \rVert$; and \emph{Space} (per-joint \emph{Directness}) as $\big(\sum_t \lVert P(t)-P(t-w)\rVert\big) / \lVert P(T)-P(t_1)\rVert$, where the lag $w$ is the $55$-frame window. Four global Effort values aggregate the per-joint factors with extremity weighting (wrists and ankles $\times 1.5$, others $\times 1.0$, after the mmpose keypoint weights). Space additionally contains five dispersion distances (head and wrists relative to the upper spine, ankles relative to the pelvis) and the pelvis trajectory's path length, net displacement, and their ratio (curvature). Shape is the convex-hull volume of the 24 joints. Body adds six inter-joint distances (wrist--shoulder L/R, ankle--knee L/R, wrist--wrist, ankle--ankle), six per-joint initiation features---each the fraction of frames whose forward windowed speed exceeds $\varepsilon$, the per-sequence standard deviation of that speed---and six inter-joint angles (left/right arm at the shoulder, the shoulder line at the pelvis, left/right knee, and the hip line at the pelvis), following Turab et al.'s ``distance and angles between the hands, shoulders, pelvis, knees, and ankles.'' Each fragment's $(T\times 61)$ matrix is reduced to a $122$-d vector by per-feature mean and standard deviation.
\begin{table}[t]
  \centering
  \small 
  \setlength{\tabcolsep}{8pt} 
  \caption{The 61 LMA descriptors by Laban category.}
  \label{tab:features}
  \begin{tabular}{l c p{5.25cm}} 
    \toprule
    Category & \# & Contents \\
    \midrule
    Effort     & 34 & velocity/Weight/Time/Flow/Space $\times$ 6 joints $+$ 4 globals \\
    Space      & 8  & 5 dispersions $+$ path, displacement, curvature \\
    Body       & 18 & 6 inter-joint distances $+$ 6 initiation $+$ 6 inter-joint angles \\
    Shape      & 1  & convex-hull volume of the 24 joints \\
    \midrule
    \textbf{Total} & \textbf{61} & \\
    \bottomrule
  \end{tabular}
\end{table}
\paragraph{Classification protocol.}
All runs balance the classes by sampling to the smallest tier ($1{,}027$ fragments per class; Tier~0 is the binding cap), drawn without replacement at random seed $42$. We report logistic regression (\texttt{max\_iter}$=3000$) and random forests ($400$ trees); the reported accuracy is the accuracy of out-of-fold predictions aggregated over the folds. Features are standardized per fold, with the scaler fit on the training split only. We evaluate under two five-fold schemes: ordinary stratified $k$-fold (\texttt{StratifiedKFold}), which yields the naive numbers, and source-video-grouped $k$-fold (\texttt{StratifiedGroupKFold}), in which every fragment of a given source video is confined to one fold; the group key is the originating video. We treat the grouped numbers as primary.

\section{Faithfulness of the LMA reimplementation}
The same LMA feature extractor appears to have been used for both dance classification \cite{turab2025dance} and emotion recognition \cite{turab2025emotion}.\footnote{The latter paper reports 54 features, but it is not clear from the text which feature was excluded.} We did \emph{not} have access to the official code for said LMA feature extractor; our extractor is a best-effort reconstruction from their papers. The papers specify the descriptor's structure and several equations, but publish neither the literal list of 55 feature names nor a per-category breakdown, and release no code. For transparency and reproducibility we record which aspects are fixed by the paper and which are our own choices.

\paragraph{Fixed by the paper.}
The following are stated in the text: the total of 55 features grouped into four Laban components (Body, Effort, Space, Shape); the six key joints (head, pelvis, both wrists, both ankles); the four Effort factors (Space, Weight, Time, Flow) with their core equations: Weight as kinetic energy, Time as acceleration magnitude, and Space as a windowed path-to-displacement ratio; the per-joint initiation event thresholded at a per-sequence standard deviation; and floor-relative body normalization. The extremity weighting ($\times1.5$ for wrists and ankles) is lifted verbatim from the mmpose keypoint-weight array the paper cites, and is likewise treated as fixed.

\paragraph{Assumed by us.}
Where the paper is silent we made explicit choices. Among these, the higher-risk ones are: \emph{(i)} the literal instantiation of the 61 names from the count-and-scaffolding budget (Table~\ref{tab:features}); \emph{(ii)} \emph{Flow} computed as mean jerk magnitude---the paper names Flow but quotes no equation, and jerk is the standard surrogate in the LMA literature; \emph{(iii)} the specific five dispersion pairs (head and wrists to the upper spine, ankles to the pelvis) and six inter-joint Body distances (wrist--shoulder, ankle--knee, wrist--wrist, ankle--ankle); \emph{(iv)} the lag $w$ in the Space and Initiation equations, which the paper leaves only as a ``short window''; and \emph{(v)} an SMPL 24-joint topology (Turab et al. may have used a different skeleton, though the six key-joint indices are unambiguous regardless). Of these, Flow-as-jerk is the least constrained by the paper, whereas the window length, the convex-hull shape proxy, the per-joint velocities, the initiation count, and the extremity weighting are the most secure. Because a literal reading of the features results in a feature count of 61---not 55---we have ultimately decided to keep all 61 features instead of dropping six features in order to force the feature count to match. For $w$, we make the assumption that it equates to the $55$-frame operating sliding window mentioned in the paper. Our reimplementation of the LMA extractor is available in a separate GitHub repository.\footnote{\texttt{https://github.com/zaiisao/dance-style-recognition}}

\section{Tier-2 dataset construction}
\label{app:tier2}
Unlike the other tiers, Tier~2 (suggestive) is not drawn from a public benchmark; we assembled it by targeted web crawling and document its construction here. Acquisition used seven YouTube search categories and four TikTok creator-channel categories (Table~\ref{tab:tier2}); the YouTube crawl was subsequently broadened with additional related queries. Because per-clip category labels were not retained in the final feature set, we report the acquisition taxonomy and platform-level totals---108 source videos yielding ${\sim}1{,}715$ fragments---rather than per-category fragment counts. All clips were manually filtered to single-performer footage and de-duplicated by source video. We do not publish per-clip links (the content is adult-adjacent and platform URLs are unstable); the queries and channels below are sufficient to reconstruct comparable data.

\begin{table}[t]
  \centering
  \caption{Tier-2 acquisition categories. YouTube was crawled by search query, TikTok by creator channel.}
  \label{tab:tier2}
  \small
  \begin{tabular}{l l}
    \toprule
    Category & Example query / channel \\
    \midrule
    \multicolumn{2}{l}{\textit{YouTube search categories (7)}} \\
    \quad Twerk              & ``twerk tutorial dance short'' \\
    \quad Reggaeton / perreo & ``reggaeton perreo dance solo'' \\
    \quad Sensual / fancam   & ``sensual dance choreography'' \\
    \quad Streamer dance cover & ``Korean streamer dance fancam'' \\
    \quad Chair dance        & ``chair dance routine'' \\
    \quad Heels              & ``heels dance choreography class'' \\
    \quad Belly dance        & ``belly dance hip isolation'' \\
    \midrule
    \multicolumn{2}{l}{\textit{TikTok creator-channel categories (4)}} \\
    \quad Twerk              & \texttt{@b00tybyjacks}, \texttt{@danceemporiumbylh} \\
    \quad Perreo / reggaeton & \texttt{@alessandra\_xsx}, \texttt{@djosocity} \\
    \quad Heels / floorwork  & \texttt{@exoticdanceacademy}, \texttt{@kheannawalker} \\
    \quad Dancehall / whine  & \texttt{@yvngmik.e2}, \texttt{@roseylucci} \\
    \bottomrule
  \end{tabular}
\end{table}

\end{document}